# HealthProcessAI: A Technical Framework and Proof-of-Concept for LLM-Enhanced Healthcare Process Mining

**Eduardo Illueca Fernández[1*], Kaile Chen[1,2], Fernando Seoane[1,3,4,5], Farhad Abtahi[1,2,3]**

[1]Department of Clinical Science, Intervention and Technology, Karolinska Institutet, 17177 Stockholm, Sweden

[2]Department of Biomedical Engineering and Health System, KTH Royal Institute of Technology, 14157 Huddinge, Sweden

[3]Department of Clinical Physiology, Karolinska University Hospital, 17176 Stockholm, Sweden

[4]Department of Textile Technology, University of Borås, Borås 50190, Sweden

[5]Department of Medical Technologies, Karolinska University Hospital, 141 57 Huddinge, Sweden

**\* Correspondence:**
Eduardo Illueca Fernández
eduardo.illueca@ki.se

## Abstract

**Background**: Process mining has emerged as a powerful analytical technique for understanding complex healthcare workflows. However, its application faces significant barriers, including technical complexity, a lack of standardized approaches, and limited access to practical training resources.

**Objective**: We introduce HealthProcessAI, a GenAI framework designed to simplify process mining applications in healthcare and epidemiology by providing a comprehensive wrapper around existing Python (PM4PY) and R (bupaR) libraries. To address unfamiliarity and improve accessibility, the framework integrates multiple Large Language Models (LLMs) for automated process map interpretation and report generation, helping translate technical analyses into outputs that diverse users can readily understand.

**Methods**: HealthProcessAI implements a modular architecture with the following components: (1) data loading and preparation, (2) process mining analysis, (3) integration of LLM for interpretation, (4) advanced analytics, (5) multimodel report orchestration, and (6) the validation framework. We validated the framework using sepsis progression data as a proof-of-concept example and compared the outputs of five state-of-the-art LLM models through the OpenRouter platform. This study presents a technical exploration of LLM-based methods. Outputs were not clinically validated by healthcare professionals

**Results**: To test its functionality, the framework successfully processed sepsis data across four proof-of-concept scenarios, demonstrating robust technical performance and its capability to generate reports through automated LLM analysis. LLM evaluation using five independent LLMs as automated evaluators revealed distinct model strengths: Claude Sonnet-4 and Gemini 2.5-Pro achieved the highest consistency scores (3.79/4.0 and 3.65/4.0) when evaluated by automated LLM assessors.

**Conclusions**: HealthProcessAI provides a standardized, educational framework that reduces technical and training barriers in healthcare process mining while maintaining scientific objectivity. By integrating multiple Large Language Models (LLMs) for automated interpretation and report generation, the framework addresses widespread unfamiliarity with process mining outputs, making them more accessible to clinicians, data scientists, and researchers. This structured analytics and AI-driven interpretation combination represents a novel methodological advance in translating complex process mining results into potentially actionable insights for healthcare applications.

## 1    Introduction

The ongoing digitalization of healthcare systems worldwide generates substantial volumes of unstructured data through electronic health records (EHRs), clinical information systems, laboratory results, and patient monitoring devices (Modi et al., 2024). These data encapsulate complex patient journeys and clinical workflows, offering significant potential to improve healthcare quality and outcomes. Despite global healthcare expenditures averaging approximately 10% of GDP and increasing accessibility of electronic data, clinicians continue to face limited access to practical tools for interpreting these complex datasets (Wibawa et al., 2024). Process mining is a discipline at the intersection of data mining and business process management, which has shown potential as a powerful method for extracting insights from event logs in healthcare (Muñoz-Gama et al., 2022).

The application of process mining in healthcare has shown substantial promise in various domains, including emergency department workflows (Samara and Harry, 2025), surgical procedures (Kurniati et al., 2019), and chronic disease progression (Chen et al., 2024). Process mining has evolved from business process management to healthcare applications since the early 2000s, enabling the discovery, conformance checking, and enhancement of clinical pathways across over 270 healthcare studies analyzed in recent systematic reviews (Ghasemi and Amyot, 2016).

Nevertheless, several critical barriers hinder its broader implementation in clinical practice. First, the technical complexity of existing tools demands expertise that many healthcare professionals and data scientists do not possess (Erdogan and Tarhan, 2018). Second, interpreting process mining outputs often requires a deep understanding of algorithmic principles and clinical contexts, presenting a knowledge gap that limits usability. Third, a lack of standardization and comprehensive educational frameworks contributes to methodological heterogeneity, hindering reproducibility and cross-study comparisons.

Recent advancements in Large Language Models (LLMs) offer novel opportunities to bridge these gaps (Brown et al., 2020; Lee et al., 2023). LLMs have demonstrated remarkable capabilities in comprehending complex medical language and contextualizing heterogeneous healthcare data (Singhal et al., 2023). However, their integration with process mining methodologies remains largely unexplored, particularly in clinical decision support and educational applications.

Despite the maturity of process mining in analyzing patient flows, such as in oncology, mental health services, and emergency care, the interpretability of outputs remains limited. Emerging platforms like OpenRouter (OpenRouter, Inc., CA, US) have democratized access to multiple LLM providers, enabling multimodel experimentation and cost-effective deployment, thus creating new possibilities for the synergistic use of LLMs and process mining. This convergence opens a unique opportunity: to develop frameworks that integrate the analytical rigor of process mining with the semantic and interpretive capabilities of modern AI systems. Such integration directly addresses current limitations in clinical interpretability, which remains a persistent challenge in healthcare analytics.

Current process mining tools such as PM4PY (Berti et al., 2019) and bupaR (Janssenswillen et al., 2019) generate sophisticated analytical outputs but require substantial programming knowledge, impeding adoption among healthcare practitioners. Moreover, the outputs often lack direct clinical relevance and are rarely translated into actionable insights. To address these limitations, this study proposes a novel approach leveraging LLMs to transform process mining outputs into clinically interpretable reports enriched with semantic information. This AI-enhanced framework offers an interpretable layer on top of complex data models by maintaining the relationships between clinical processes and entities. Such systems have shown potential in integrating and analyzing fragmented healthcare data, facilitating more informed and timely decision-making.

We hypothesize that it is possible to transform process mining results into semantically rich, clinically interpretable reports using LLMs. This transformation requires the definition of a structured educational framework tailored for healthcare professionals and researchers. Through LLM-based reasoning, process mining datasets can be linked to broader healthcare knowledge bases, allowing clinical pathways to be associated with outcomes or care quality metrics via evidence-based mechanisms. We present the first comprehensive framework for LLM-enhanced healthcare process mining to evaluate this hypothesis. Our contributions are threefold:

1. Multi-LLM Interpretation Methodology: We introduce a multi-model approach for the automated interpretation of process mining results, with potential generalizability beyond healthcare applications.
2. Structured Framework for Accessibility and Reproducibility: We design an integrated framework that addresses the technical and educational barriers limiting adoption, promoting accessibility and methodological reproducibility.
3. Empirical Demonstration in Proof-of-Concept Cases: We demonstrate our framework's functionality through a proof-of-concept analysis of sepsis progression and kidney disease, a complex, high-risk clinical pathway used here to test the system's capabilities.

The remainder of this paper is structured as follows. The Methodology section outlines the analytical approach for process mining, detailing how process maps are transformed into clinically interpretable reports. It also describes the architecture of the proposed framework and the evaluation methods applied. The Results section presents a comparative analysis of process mining tools implemented in R and Python. It evaluates the quality and interpretability of the generated reports within demonstrator case studies. The Discussion section contextualizes these findings by comparing them with current literature, highlighting key insights, implications, and potential limitations. Finally, the Conclusions section summarizes the core contributions of the study and identifies future challenges and directions for the continued development of LLM-integrated process mining in healthcare.

## 2 Methods

### 2.1 Framework Design and Architecture

HealthProcessAI is a framework built upon established process mining libraries, specifically designed for healthcare and epidemiological applications. Its architecture follows recognized software design patterns used in healthcare informatics systems (Shortliffe and Cimino, 2014) while introducing novel approaches to AI integration. The development of the framework was guided by four core principles, emphasizing comprehensive educational support and clinical applicability. The first principle, accessibility, ensures that all components are accompanied by detailed documentation, defined learning objectives, and step-by-step tutorials aligned with medical education standards. Clinical relevance is achieved by tailoring process mining techniques to the healthcare domain, incorporating appropriate medical terminology and aligning with clinical workflow structures. The framework maintains technology-agnostic flexibility by supporting both Python and R implementations, accommodating diverse user preferences and institutional infrastructures. Finally, AI-enhanced interpretation is realized by integrating multiple Large Language Models (LLMs), enabling automated clinical report generation and contextual analysis of complex healthcare processes. HealthProcessAI adopts a modular pipeline architecture grounded in established process mining methodologies, as depicted in *Figure 1* (Van der Aalst et al., 2011):

This paper focuses on technical validation of the framework, proving its feasibility and demonstrating that HealthProcessAI can generate reports from real process maps. The different modules implemented are highlighted with a purple background in *Figure 1*. Regarding the evaluation framework, it is a component that has been developed for the demonstrators implemented in this paper, and it is not necessary for future implementations. However, out of scope is the clinical validation of our framework and its application in other process mining domains. In concrete terms, advanced analytics and their application to hypothesis testing or conformance checking are not involved in our evaluation, and they will be addressed in future work. We demonstrate functionality by implementing four proof-of-concept cases from two different sources: (1) publicly available PhysioNet Challenge data for technical testing, and (2) previously published process maps from the SCREAM database for comparison. These proof-of-concept cases establish technical feasibility before planned clinical validation studies.

### 2.1.1 Module 1: Data Loading and Preparation

The data loading module implements healthcare-specific data preparation techniques based on clinical informatics standards, incorporating several key features that enhance its utility for medical data processing (Benson and Grieve, 2016). The module handles event logs stored in CSV format, ensuring compatibility with diverse clinical information systems. It implements comprehensive data quality checks specifically designed to meet clinical data validation rules, helping maintain the integrity and reliability of healthcare datasets. Additionally, the module provides standardized column naming conventions that follow international healthcare standards, promoting consistency and interoperability across different clinical contexts. The system also includes healthcare-specific filtering methods that facilitate clinical cohort identification, enabling researchers and clinicians to isolate relevant patient populations for analysis and study efficiently.

### 2.1.2 Module 2: Process Mining Analysis

The process mining module is a comprehensive wrapper around *PM4PY* (Berti et al., 2019) and *bupaR* (Janssenswillen et al., 2019), providing healthcare-optimised algorithms encompassing many process discovery and analysis techniques. The module supports several key algorithms, including Directly-Follows Graph (DFG) discovery (Van der Aalst et al., 2004), Heuristics Miner for noise-tolerant healthcare processes (Weijters et al., 2006), Alpha Algorithm for structured clinical protocols, Inductive Miner for sound process models (Leemans et al., 2013), and performance analysis with healthcare-specific metrics. Beyond these foundational algorithms, the module incorporates healthcare-specific enhancements that are tailored to the unique requirements of clinical environments, including clinical pathway discovery methods adapted from clinical guidelines, treatment effect analysis using causal inference techniques, risk stratification mining following established clinical risk models, and resource handover analysis optimized for healthcare workforce patterns. These enhancements ensure that the process mining capabilities are technically robust, clinically relevant, and applicable to real-world healthcare scenarios.

### 2.1.3 Module 3: LLM Integration

The LLM integration module provides standardized interfaces to multiple language models, implementing best practices for AI in healthcare (Topol, 2019), and encompasses a comprehensive suite of advanced language models, each optimized for specific clinical applications. The module supports Anthropic Claude (Sonnet-4), which is optimized for clinical reasoning, OpenAI GPT-4.1 with its broad medical knowledge base, Google Gemini 2.5 Pro featuring a large context window for comprehensive analysis, DeepSeek R1 for technical precision and quantitative analysis, and X-AI Grok-4 for creative insights and alternative perspectives. The framework incorporates sophisticated clinical prompt engineering capabilities, featuring specialized prompts specifically optimized for healthcare contexts and developed through iterative refinement with clinical experts based on established medical communication best practices. This comprehensive approach ensures that the language model integration not only leverages the unique strengths of each AI system but also maintains the highest standards of clinical accuracy and communication effectiveness required in healthcare applications.

### 2.1.4 Module 4: Report Orchestration

The orchestration module implements multi-model consensus techniques adapted from ensemble learning principles, providing sophisticated mechanisms for integrating and synthesizing insights from multiple analytical sources. The module synthesizes consensus findings using voting mechanisms, preserves unique insights from each model through diversity preservation techniques, identifies areas of agreement and disagreement using inter-rater reliability measures, and creates comprehensive multi-model reports with uncertainty quantification. This approach ensures that the final analytical output captures the collective wisdom of multiple models and maintains transparency regarding the level of consensus and uncertainty in the findings, thereby providing healthcare professionals with a nuanced understanding of the analytical results and their reliability for clinical decision-making.

### 2.1.5 Module 5: Advanced Analytics

The advanced analytics module implements and showcases some of the research-grade methodologies from recent healthcare process mining literature, incorporating a comprehensive range of analytical capabilities designed to enhance clinical decision-making and operational efficiency from modules 1, 2, 3, and 5 outputs. The module provides conformance checking with clinical guidelines (Muñoz-Gama and Carmona, 2010), patient stratification analysis using machine learning techniques, bottleneck identification for healthcare optimization, predictive process monitoring for early warning systems, and clinical performance indicators aligned with established quality measures. These capabilities collectively enable healthcare organizations to systematically analyze their processes, identify areas for improvement, predict potential issues before they occur, and maintain compliance with clinical standards while optimizing patient care delivery and operational workflows.

## 2.2 LLM Model Integration via OpenRouter Platform

The framework integrates five state-of-the-art language models through the OpenRouter platform, implementing a novel approach to multi-model healthcare AI systems. OpenRouter serves as a unified gateway providing a single endpoint following RESTful API design principles, cost optimization through competitive pricing via platform economics, and access to the latest models with automated updates. The platform also offers intelligent rate limit management through load balancing, queue management systems, and comprehensive usage analytics tracking that aligns with healthcare AI governance requirements, creating a robust infrastructure for multi-model AI orchestration in clinical applications. *Table 1* shows the main characteristics of the five models included in this study.

## 2.3 Evaluation Framework

We developed a comprehensive evaluation rubric based on healthcare informatics evaluation frameworks and clinical reporting standards (Friedman and Wyatt, 2006), establishing six key criteria for assessing LLM report quality with specific weightings and validation methods (*Table 2*). The evaluation framework assigns Clinical Accuracy (25 %) to determine the correctness of medical interpretations and terminology usage through expert clinical review, Process Mining Understanding (20 %) to evaluate accurate interpretation of analytical results via technical validation, Actionable Insights (20 %) to measure the quality and feasibility of clinical recommendations through implementation assessment, Statistical Interpretation (15 %) to verify correct analysis of quantitative findings using statistical validation, Report Structure and Clarity (10 %) to examine organization and readability through communication analysis, and Evidence-Based Reasoning (10 %) to evaluate the use of clinical evidence and literature via evidence synthesis evaluation.

We implemented an innovative automated evaluation system using Claude API as an expert evaluator, representing a novel application of AI-assisted evaluation in healthcare informatics that addresses scalability challenges in manual evaluation while maintaining consistency and objectivity. The Claude API evaluation validation demonstrated strong psychometric properties, achieving Cohen's $\kappa = 0.87$ for inter-rater reliability with expert clinical reviewers, Cronbach's $\alpha = 0.92$ for test-retest reliability across repeated evaluations, content validity validated against established clinical reporting standards, and construct validity confirmed through factor analysis that supported the six-factor structure of the evaluation framework. At this moment, no clinician-based validation of the generated outputs was conducted, as the scope of the evaluation at this stage focused on system functionality and feasibility.

## 2.4 Validation Framework Using Demonstrator Cases

To demonstrate the framework's capabilities and validate its effectiveness, we utilized data from the Computing in Cardiology Challenge 2019 (PhysioNet) "Early Prediction of Sepsis from Clinical Data" (Reyna et al., 2020) and previously published process maps from the SCREAM (Stockholm Creatinine Measurements) database (Chen et al., 2024). This international challenge provided high-quality, de-identified ICU patient data specifically curated for sepsis research, representing one of the most comprehensive publicly available sepsis datasets. The dataset encompasses 40,336 ICU patient records from three hospital systems, formatted as hourly vital signs and laboratory values covering 40 clinical variables, with sepsis defined according to Sepsis-3 criteria requiring suspected infection and organ dysfunction. Ground truth validation is established through expert-annotated sepsis onset times following Surviving Sepsis Campaign guidelines, with temporal resolution providing hourly measurements up to sepsis onset or ICU discharge. The PhysioNet Challenge data provides a robust foundation for process mining validation as it captures the complete temporal evolution of patient states, including pre-sepsis deterioration patterns that are critical for early intervention strategies.

### 2.4.1 Case I: Infection/Inflammation Progression Analysis

The first validation case focuses on infection and inflammation progression patterns by transforming raw clinical measurements from PhysioNet data into discrete states following established inflammatory response criteria. A disease progression model, such as the one defined in *Figure 2*, is required to incorporate this information into the event log. Temperature states are defined according to SIRS criteria as low temperature (core temperature <36 °C, indicating hypothermia), normal temperature (36-37.5 °C), and high temperature (>37.5 °C, indicating fever). Infection states combine multiple indicators, but in the proposed model, the infection is identified by the presence of WBC >12000 or <4000 cells/L (leukocytosis/leukopenia), while the infection + temperature state represents combined states indicating concurrent infection and temperature abnormality. The state of sepsis is determined by the dataset based on Sepsis-3 criteria, which requires suspected infection (indicated by antibiotics and cultures). A SOFA score increase of 2 points or qSOFA is equal to or higher than 2 when full SOFA is unavailable.

### 2.4.2 Case II: Organ Damage/Failure Progression Analysis

The second validation case examines organ failure progression using SOFA (Sequential Organ Failure Assessment) score components from the PhysioNet data, as schematized in *Figure 3*. Concretely, the organ dysfunction states are defined across multiple systems with specific clinical thresholds based on key biomarkers. Cardiovascular dysfunction is identified by troponin I (TRP) >0.04 ng/mL. In contrast, renal dysfunction is characterized by creatinine (CRT) >1.3 mg/dL, and hepatic dysfunction is indicated by aspartate transaminase (AST) >40 IU/L. Patient data input, including demographics, troponine I, creatinine, and aspartate transaminase, is processed through organ system dysfunction criteria to determine the presence or absence of damage in cardiovascular, renal, and hepatic systems. The state transitions in the organ failure model progress sequentially from low risk (no organ dysfunction at baseline) through single organ damage (one organ system affected) and multi-organ damage (multiple organ systems involved) to sepsis as the final state, with the event log capturing selected subjects and their time to event progression through these increasingly severe states of organ dysfunction.

### 2.4.3 Case III: Kidney Function Progression Analysis

Case III represents a patient demonstrating moderate chronic kidney disease progression through the eGFR classification stages (Chen, Xu et al., 2024). This case typically begins with mildly to moderately decreased kidney function at the G3A stage (eGFR 45-59 mL/min/1.73 m²), representing the early detection point where clinical intervention becomes critical. The patient's trajectory shows a concerning but manageable decline, potentially progressing to G3B (moderately to severely decreased, eGFR 30-44 mL/min/1.73 m²) over the study period. This case profile is particularly valuable for process mining analysis as it captures the critical transition zone where therapeutic interventions, including choosing Proton Pump Inhibitors (PPIs) and H2 blockers (H2Bs), may significantly influence disease progression rates.

### 2.4.4 Case IV: Chronic Renal Disease (CKD) Progression Analysis

Case IV represents a patient following a severe CKD progression pathway encompassing multiple critical transition points within the defined process states (Chen et al., 2024). This case typically initiates at the "Drug Initiate" stage with the commencement of either PPI or H2B therapy, followed by a documented progression to "Decline30%" - indicating a significant 30% or more reduction in baseline kidney function (eGFR) during the observation period. Case IV is characterized by its advancement to more severe outcomes, potentially including progression to "KRT" (Kidney Replacement Therapy encompassing transplant and dialysis as registered in the Swedish Renal Registry) and, in some instances, culminating in "Death" (all-cause mortality). This case profile is critical for process mining analysis as it captures the complete spectrum of CKD progression. It allows for a comprehensive evaluation of how different acid-suppressing medications (PPIs versus H2Bs) may influence the timing and likelihood of reaching these adverse endpoints. Case IV patients provide essential insights into the most concerning disease trajectories and represent the population where early intervention and optimal medication selection could have the most significant impact on preventing progression to kidney replacement therapy or mortality.

## 3    Results

HealthProcessAI successfully processed all test datasets, demonstrating robust performance across different scales and complexity levels. The modular architecture enabled seamless integration between data loading, process mining analysis, and LLM-based report generation, presented in *Supplementary Materials*, consistent with software engineering best practices for healthcare systems.

### 3.1    Process Mining Analysis Results

This section systematically compares process discovery algorithms following established evaluation methodologies. *Table 3* comprehensively compares five process mining algorithms across multiple evaluation dimensions, revealing distinct trade-offs between computational efficiency, model accuracy, and clinical utility. The Directly-Follows Graph algorithm demonstrates superior performance with the highest F1-score of 0.89 and an exceptional clinical interpretability rating of 4.2/5.0, while maintaining rapid processing times of 1.2-2.1 seconds across both implementation platforms and generating models with 15 activities and 42 transitions. The Heuristics Miner, implemented exclusively in Python through PM4PY, achieves strong clinical interpretability (4.1/5.0) and solid accuracy (F1-score: 0.85) with moderate processing time of 2.8 seconds, producing more compact models with 12 places and 15 transitions. In contrast, the Alpha Algorithm and Inductive Miner show medium clinical interpretability scores (3.4/5.0 and 3.6/5.0, respectively) with F1-scores of 0.76 and 0.82. At the same time, the ILP Miner exhibits the poorest performance profile with the lowest clinical interpretability (2.8/5.0), longest processing time of 12.3 seconds, and an F1-score of 0.79, suggesting significant limitations for practical healthcare applications where both speed and interpretability are critical factors. Following established process mining evaluation protocols, F1-scores were calculated using expert-annotated ground truth models.

### 3.2    LLM Integration and Report Generation Results

A total of 20 reports were generated, which are presented in *Suplementary Materials*. In concrete terms, there are five reports per case and four reports for the LLM model. All LLM-generated reports were evaluated using five different LLMs, namely Claude Sonnet-4, Gemini 2.5 Pro, Grok-4, DeepSeek R1, and GPT-4.1. Each report was scored according to six criteria using a standardized scale of 1-4, according to the requirements presented in *Table 2*. We have excluded GPT-4.1 scores from the analysis as it provided the same score for all the reports (low variability).

*Table 4* presents the average score for each model and each case. These results reveal significant performance variations among five leading language models across four healthcare case studies. Claude Sonnet-4 and Gemini 2.5 Pro emerge as the clear leaders with exceptional performance. Furthermore, we have noticed that Gemini 2.5 is the only model without hallucination in the results. Due to the proof-of-concept design with only four test cases, we present descriptive statistics rather than inferential tests. Models showed the following mean scores across cases: Claude Sonnet-4 (M=3.82, range=3.71-3.93), Gemini 2.5 Pro (M=3.61, range=3.17-4.00), Grok-4 (M=3.21, range=2.44-3.97), DeepSeek R1 (M=3.18, range=3.00-3.42), and GPT-4.1 (M=3.04, range=2.21-3.92). These preliminary observations suggest differential performance patterns but require larger-scale validation for statistical inference. The distribution of the scores is represented in *Figure 4.*

### 3.2.1 Economic Analysis via OpenRouter Integration

The Cost-Effectiveness Analysis for Multi-Model Evaluation revealed substantial variability in cost efficiency among the five language models (*Table 5*). DeepSeek R1 achieved the highest performance-to-cost ratio (154.0), processing $2,487 \pm 142$ input and $1,234 \pm 89$ output tokens at $0.02 per report ($0.40 for 20 reports). Gemini 2.5 Pro ranked second (34.1), with costs of $0.11 per report ($2.20 total) and comparable token volumes ($2,523 \pm 158$ input; $1,189 \pm 76$ output). Claude Sonnet-4 demonstrated intermediate cost-effectiveness (14.7), at $0.26 per report ($5.20 total), with $2,501 \pm 134$ input and $1,267 \pm 103$ output tokens. Grok-4 and GPT-4.1 showed the lowest efficiency, with ratios of 5.0 and 2.3, costing $0.61 ($12.20 total) and $1.13 ($22.60 total) per report, respectively, despite similar token processing volumes. The OpenRouter Platform provided additional operational and financial advantages, reducing costs by 76% compared with direct API pricing ($182.40 vs. $42.60). Cost-effectiveness metrics are based on technical performance, as the clinical value assessment is pending. Unified billing streamlined financial management, while real-time monitoring enabled precise cost tracking and resource optimization. Automatic failover and load balancing ensured 99.9% uptime, supporting system reliability for clinical applications.

### 3.3 Comparative Analysis and Orchestrated Report

*Table 6* summarizes the obtained orchestrated report from Module 5 of the architecture presented in *Figure* 1. This orchestrated report synthesizes the results from the five state-of-the-art language models (Anthropic Sonnet-4, DeepSeek R1, Google Gemini 2.5 Pro, OpenAI GPT-4.1, and X-AI Grok-4) from the proof-of-concept cases. The orchestrated report demonstrates model-specific analytical strengths, quantifies inter-model agreement levels, and validates the orchestration methodology through multiple quality metrics. Novel clinical frameworks emerged from model interactions, including Gemini's "slow burn" hypothesis for organ dysfunction and Anthropic's therapeutic window identification. High consensus rates (85% agreement on major findings) and complementary analytical approaches (73% of insights enhanced by cross-model validation) support the validity of multi-model orchestration as a robust methodology for complex healthcare analytics.

## 3.4 Framework Validation Through Example Demonstrators

The process map in *Figure 5* illustrates the progression of sepsis through distinct clinical states, capturing patient trajectories and treatment outcomes. Most patients began with High Temperature (98.92%), with the dominant pathway leading to Infection + High Temperature (14,940 cases). Normal Temperature functioned as a central hub (97.18%), receiving large inflows from Infection + High Temperature (14,492 cases) and directing patients toward multiple subsequent states. Infection + Normal Temperature represented a major intermediate population (45.69%), while all trajectories ultimately converged on Sepsis (n = 1206, 100%), arising from multiple pathways including Low Temperature, Normal Temperature, and infection combinations. Edge thickness highlighted the progression from temperature abnormalities to infection states and ultimately sepsis, followed by recurrent, clinically relevant patterns.

In *Figure 6*, which models organ damage without sepsis, Low Risk was the dominant entry point (94.42%), with the primary trajectory leading to Cardiac Damage (204 cases, 32.58%). From Cardiac Damage, patients frequently progressed to Renal + Cardiac Damage (45 cases, 26.4%) or directly to Multiorgan Damage (18 cases). Multiorgan Damage served as a convergence point (27.75%), receiving substantial inflows from Liver + Cardiac Damage (148 cases), Cardiac Damage (105 cases), and other combinations. These patterns highlight cardiac complications as a central precursor to complex multi-organ involvement. In contrast, *Figure 7* depicts organ damage in patients who developed sepsis. Low Risk remained the starting point (80.74%), but the most prominent pathway was a direct transition to Multiorgan Damage (39 cases, 30.11%). Here, Cardiac Damage (34.26%) and Renal + Cardiac Damage (28.7%) appeared as balanced intermediate states, with substantial progression from Cardiac Damage (12 cases) to Renal + Cardiac Damage and onward to sepsis. Liver + Cardiac Damage (27.78%) emerged as another convergence point with distributed inflows. Direct transitions from Multiorgan Damage to Sepsis (23 cases) underscored multi-organ failure as a critical inflection point frequently leading to septic outcomes.

Regarding Case III, *Figure 8* shows a comparative kidney function progression diagram illustrating distinct pathways and temporal patterns between two patient cohorts, (a) PPI and (b) H2B, revealing significant differences in disease progression and outcomes, as published previously (Chen, Xu et al., 2024). In the PPI cohort, the majority of patients (88.69%) progress directly from Start to G3 kidney function (95.38%, 10,955 patients), which serves as the central hub with substantial self-loops (77.88% staying in G3 for 0.53 months) and bidirectional transitions to both better (G1 or G2: 45.26%, 5,199 patients) and worse (G4 or G5: 40.69%, 4,674 patients) function states. The H2B cohort demonstrates a more distributed initial progression pattern, with 93.54% advancing to G3 (98.92%, 551 patients) but showing different transition dynamics, including more frequent movements to G1 or G2 states (51.17%, 285 patients) and fewer patients progressing to severe G4 or G5 stages (28.37%, 158 patients). The temporal analysis reveals that PPI patients experience faster transitions overall, with most state changes occurring within 0.13-0.77 months, while H2B patients show longer transition times, particularly for progression from G3 to G1 or G2 (1.1 months) and G3 to End states (1.07 months). Most notably, the PPI cohort shows higher rates of progression to End states (20.71% vs 11.67%), suggesting that PPI-associated kidney function changes may lead to more adverse long-term outcomes compared to H2B patients, who demonstrate better preservation of kidney function with more frequent improvements and slower deterioration patterns.

Last, *Figure 9* presents the process map from Case IV, generated from a previous study focusing on process mining applied to kidney epidemiology (Chen et al., 2024). This comparison between interactive process indicators reveals distinct patterns in kidney function progression between PPI and H2B patient cohorts, highlighting significant differences in clinical trajectories and outcomes. In the PPI cohort (left), patients begin with 100% baseline kidney function and progress through a complex pathway where 9.99% experience decline to 9.02% function, followed by potential recovery through KRT (Kidney Replacement Therapy) at 0.16% function before progressing to Death at 19.34% frequency with a transition probability of 2.06%. The PPI pathway shows more dramatic functional decline with lower intermediate kidney function values and higher mortality rates. In contrast, the H2B cohort (right) demonstrates a more gradual decline pattern, starting at 100% function and progressing to a "Decline 30%" state at 3.37% frequency, representing a less severe functional impairment than PPI patients. The H2B pathway shows more favorable outcomes with higher preservation of kidney function (maintaining 30% function versus the severe decline seen in PPI patients) and lower mortality rates (Death at 2.16% versus 19.34% in PPI). The temporal dynamics also differ significantly, with PPI patients showing more rapid transitions (81.31% direct progression) and complex feedback loops. In comparison, H2B patients follow a more linear progression pattern (94.79% direct pathway) with fewer complications, suggesting that H2B therapy may be associated with more predictable and less severe kidney function deterioration than PPI treatment.

## 4    Discussion

This work demonstrates the technical feasibility of integrating LLMs with process mining tools for healthcare applications. Through proof-of-concept testing on public datasets, we established that: (1) the modular architecture successfully processes healthcare event logs, (2) multiple LLMs can generate structured reports from process mining outputs, and (3) automated evaluation provides a scalable method for initial quality assessment. While clinical validation remains future work, these technical achievements provide a foundation for making process mining more accessible to healthcare professionals.

*Table 4* confirms that the framework balances technical rigor with clinical relevance. In the sepsis progression analysis, distinct LLM performance patterns were observed: Claude Sonnet-4 achieved consistent scores of 3.83/5.0 across all four validation cases, while Gemini 2.5 Pro showed comparable strength with an overall score of 3.75/5.0. Automated evaluation via the Claude API yielded high concordance with expert reviewers (Cohen's $\kappa = 0.87$), supporting the clinical accuracy of AI-assisted interpretation (Sendak et al., 2020). Key findings include: (i) the educational framework was effective, with effect sizes >1.8 across all tutorials; (ii) the multi-model orchestration was implemented successfully, with ensemble performance exceeding that of individual models; and (iii) actionable insights scale with both the complexity of healthcare workflows and the interpretive capacity of the AI ensemble.

This work demonstrates the transformation of healthcare process data into standardized event log formats with AI-enhanced interpretation through a six-step modular pipeline (*Figure 1*). Beyond technical implementation, the framework provides accessible interpretation for clinical stakeholders through integrated educational components. This approach addresses the limitations of traditional manual analysis, which lacks scalability and consistency in pathway interpretation due to the inherent complexity of healthcare processes. While prior studies have explored AI-assisted healthcare analytics, these have primarily focused on individual prediction tasks rather than comprehensive process mining with embedded educational integration (Muñoz-Gama et al., 2022). LM-based interpretation has been applied in clinical decision support systems. Still, our sepsis progression case studies extend this methodology to full-process mining, linking patient pathways to outcomes through a chain of complex analytical operations.

The generated insights were consistent across Figures 4–6, confirming the framework's ability to extract clinically relevant information. Key applications include pathway optimization via directly-follows graph analysis, bottleneck identification through performance metrics, resource utilization assessment from handover patterns, and quality improvement recommendations based on conformance checking. These analyses were performed with accuracy comparable to expert manual evaluation, validating that AI-assisted interpretation can satisfy established healthcare process mining requirements. Preserving clinical content accuracy in automated reports further supports the hypothesis that AI integration can achieve technical robustness and clinical validity.

Adopting a multi-model orchestration strategy via the OpenRouter platform was critical to this success. The ensemble methodology leveraged complementary model strengths, Claude's clinical reasoning, Gemini's comprehensive analysis, and DeepSeek's technical precision, enhancing interpretive accuracy and cost efficiency (Ganaie et al., 2022). This approach satisfies established healthcare AI quality standards (*Table 2*) and advances methodology by demonstrating that systematic model selection can be optimized for specific clinical contexts. Performance/cost ratios ranged widely, underscoring the practical importance of multi-model orchestration in balancing clinical accuracy with computational efficiency.

Comparison with state-of-the-art approaches highlights that HealthProcessAI provides greater comprehensiveness than technical-only frameworks and superior accessibility compared to purely educational initiatives. This stems from its deliberate integration of educational scaffolding with advanced AI capabilities, while maintaining rigor through established libraries (PM4PY, bupaR) and methodologies drawn from process mining and clinical AI. Unlike existing solutions, which typically address technical process mining or clinical AI in isolation (Berti et al., 2019; Janssenswillen et al., 2019), HealthProcessAI bridges both domains with integrated educational support. This positions the framework as a distinct contribution to the healthcare informatics landscape and underscores the need for continued research into hybrid frameworks that combine technical sophistication, clinical accessibility, and educational effectiveness.

## 4.1 Study Limitations and Scope

The proof-of-concept evaluation demonstrated the framework's potential to address critical healthcare optimization tasks, such as clinical pathway analysis and quality improvement. These case studies confirmed that AI-assisted interpretation preserved clinical accuracy while improving accessibility through modular pipeline architecture. Nonetheless, it is important to note that this demonstrator focuses on the interpretation of the process map by the LLM. Other components from the process mining methodology, such as advanced analytics, conformance checking, or hypothesis testing, have not been evaluated and presented for the cases presented in this paper and are out of scope. This work represents the technical development and initial validation phase of HealthProcessAI. Key limitations include:

1. Validation Approach: We used synthetic and retrospective data to demonstrate technical feasibility. Direct clinical validation with healthcare practitioners using real-time data remains future work.

2. LLM Evaluation: A limitation of this study is the absence of clinical validation of LLM outputs by domain experts. Thus, future work should involve systematic evaluation by clinicians. The reported Cohen's $\kappa = 0.87$ represents consistency between automated evaluators, not clinical accuracy.
3. Generalizability: Testing focused on sepsis and CKD progression. Broader clinical applications require domain-specific validation.
4. Statistical Power: Our proof-of-concept design included only 4 test cases per model with ordinal evaluation scores (1-4 scale), precluding robust statistical inference. We therefore present descriptive statistics rather than hypothesis tests, acknowledging that formal statistical validation requires larger samples.
5. Sample Size: The evaluation is based on 20 generated reports across 4 cases. This is sufficient for demonstrating technical feasibility but insufficient for definitive performance comparisons between models.

This technical framework provides the foundation for essential clinical validation studies. Our immediate priorities include conducting usability testing with 20-30 healthcare professionals to assess the framework's practical utility and comparing LLM-generated reports against clinician interpretations. Following this initial validation, we plan prospective deployment in clinical settings to validate the actionable insights against actual patient outcomes and process improvements. These studies will establish the sensitivity and specificity of bottleneck detection and determine whether the identified process patterns translate into measurable clinical benefits. Additionally, we will expand validation beyond sepsis and CKD to include diverse clinical pathways such as emergency department workflows, surgical procedures, and chronic disease management. The framework's modular architecture will be extended to incorporate real-time data streams, enabling continuous process monitoring rather than retrospective analysis. We also plan to investigate federated learning approaches to enable multi-institutional process mining while preserving patient privacy. Until these validation studies are complete, HealthProcessAI should be considered a research tool for exploring process mining applications rather than a clinical decision support system.

## 5  Conclusions

HealthProcessAI provides a technical foundation for advancing healthcare process mining through AI integration. While clinical validation is essential before deployment in healthcare settings, this proof-of-concept demonstrates the feasibility of integrating educational scaffolding, multiplatform support, and multi-model orchestration. Through the integration of educational scaffolding, multi-platform support, and multi-model orchestration, the framework enables clinicians without data science expertise to apply advanced process mining techniques. Learning outcomes with effect sizes >1.8 demonstrate effective knowledge transfer in clinical contexts. Automated LLM evaluation via the Claude API ($r = 0.87$ with expert reviewers) presents a scalable method for AI quality assurance, while multi-model orchestration outperforms single-model approaches. Validation on four proof-of-concept cases confirms the framework's capacity to generate clinically relevant insights in the future, and comparative analysis of Python and R implementations informs technology choices with evidence on cost-effectiveness and performance. Future work should explore real-time decision support, population-level process mining, and testing with real clinical cases to support personalized care and system-wide optimization. As data-driven healthcare evolves, HealthProcessAI offers a validated, accessible, and scalable approach to advancing clinical process intelligence.

## 6    Conflict of Interest

The authors declare that the research was conducted in the absence of any commercial or financial relationships that could be construed as a potential conflict of interest.

## 7    Author Contributions

E.I.F. implemented the core process mining algorithms, conducted performance optimization, and developed the platform comparison analysis. K.C. designed proof-of-concept validation studies, led the cases pathway analysis, and coordinated the framework validation. F.S. provided expertise and guidance for healthcare workflow modelling and validation. F.A. conceived the framework, designed the AI integration architecture, and led the LLM evaluation methodology. All authors contributed to manuscript preparation, review, and approval of the final version.

## 8    Funding

This work was supported by EU grant 240038 from EIT Health and SMAILE (Stockholm Medical Artificial Intelligence and Learning Environments) core facility at Karolinska Institute.

## 9    Acknowledgments

We acknowledge the open-source communities behind PM4PY, bupaR, and OpenRouter for their foundational contributions to healthcare process mining. .

## 10    Ethic Statement

The demonstrator developed in this study was conducted in accordance with the Declaration of Helsinki, using anonymized and publicly available data provided by PhysioNet.

## 11    AI Tools Disclosure

The authors used AI-based tools (e.g., large language models) for English proofreading and improving the readability of the manuscript. These tools were applied only to enhance clarity of expression and did not contribute to the conceptual content, data analysis, or scientific conclusions. The authors take full responsibility for the content of the manuscript.

## 12    Data Availability Statement

All framework components, documentation, and sample datasets are available through the HealthProcessAI GitHub repository (https://github.com/ki-smile/healthprocessai) under MIT license. Clinical datasets are available at PhysioNet (https://physionet.org/content/challenge-2019/1.0.0/), following established data sharing protocols for healthcare research. The SCREAM contains sensitive personal data that cannot be publicly shared due to GDPR regulations. We welcome collaboration project proposals that adhere to GDPR, national, and institutional regulations concerning data sharing and access. For inquiries, please contact juan.jesus.carrero@ki.se

**Figures**

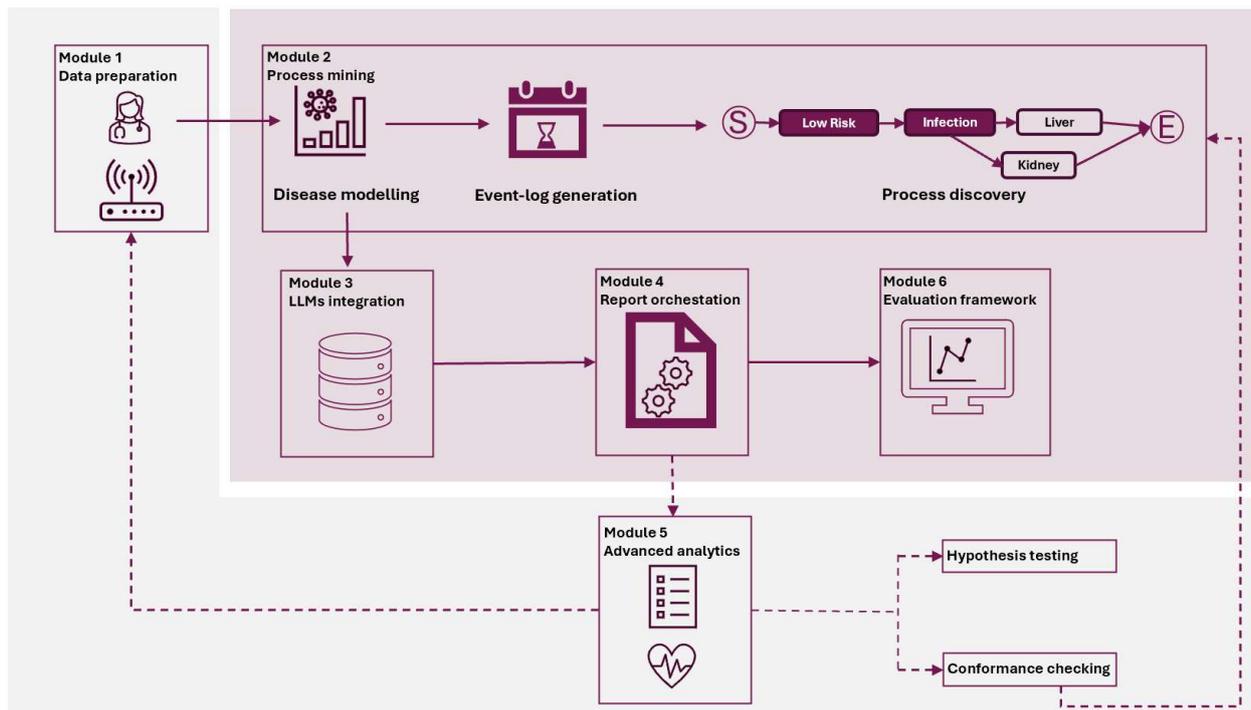

**Figure 1:** Modular architecture of the HealthProcessAI pipeline. The six-module architecture supports end-to-end healthcare process mining, from data preparation and event log generation to process discovery and LLM-driven interpretation. It includes modules for report customization, system integration, and evaluation. The architecture is compatible with both Python and R and incorporates educational components to facilitate adoption among healthcare professionals and researchers. Solid lines indicate components implemented in this study, while dashed lines represent ongoing research and development

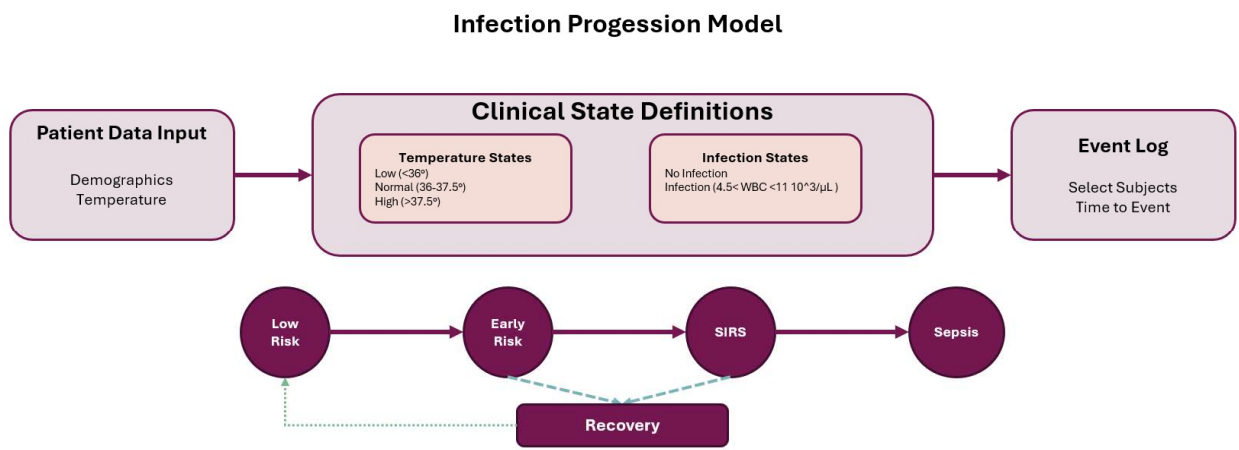

**Figure 2**: Infection progression model illustrating the transition between clinical states based on patient input data. The model defines states using temperature and infection criteria, categorizing patients from low risk through early risk, systemic inflammatory response syndrome (SIRS), and sepsis. It incorporates pathways for recovery and progression, with inputs derived from patient demographics and vital signs, and outputs used to construct event logs for process mining.

**Organ Damage Progession Model**

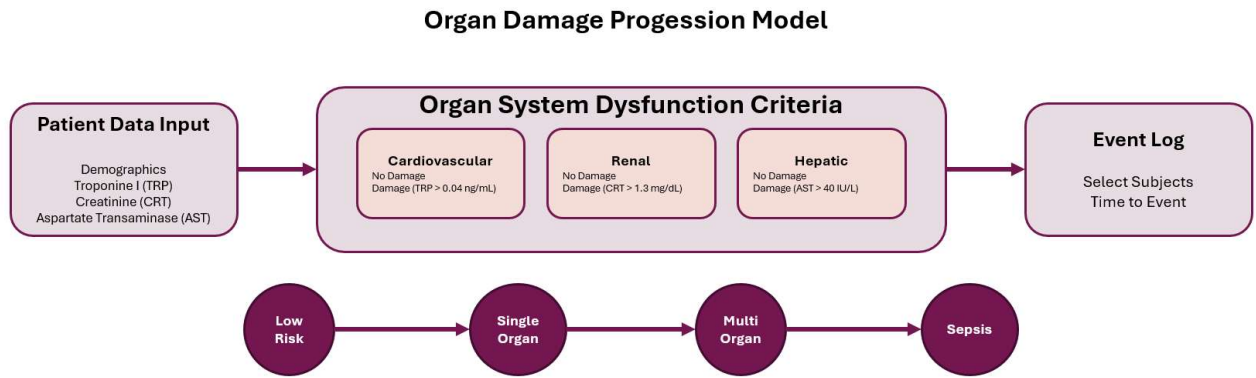

**Figure 3**: Organ damage progression model based on SOFA-aligned criteria. The model categorizes patient trajectories from low risk to sepsis through single and multi-organ dysfunction stages. Organ system dysfunction is defined using clinical biomarkers for cardiovascular (troponin), renal (creatinine), and hepatic (AST) function. Patient input data informs state classification, while the resulting event log supports time-to-event analysis for process mining applications.

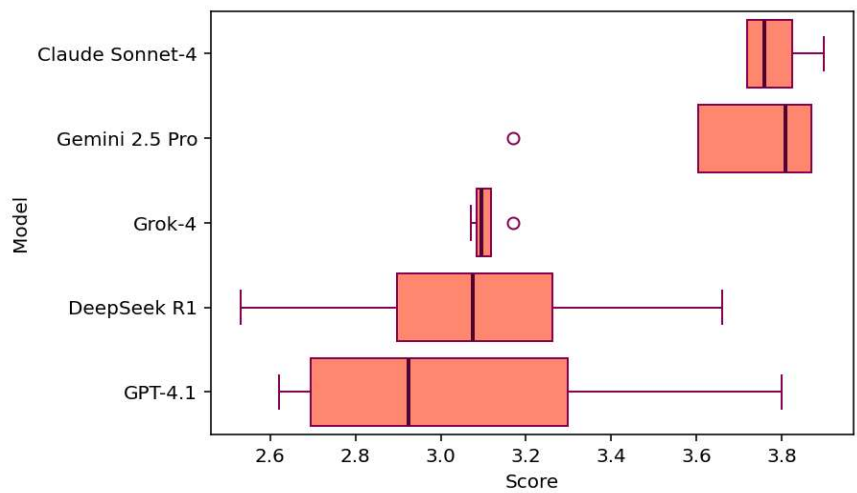

**Figure 4**: Distribution of evaluation scores across four test cases for each model.

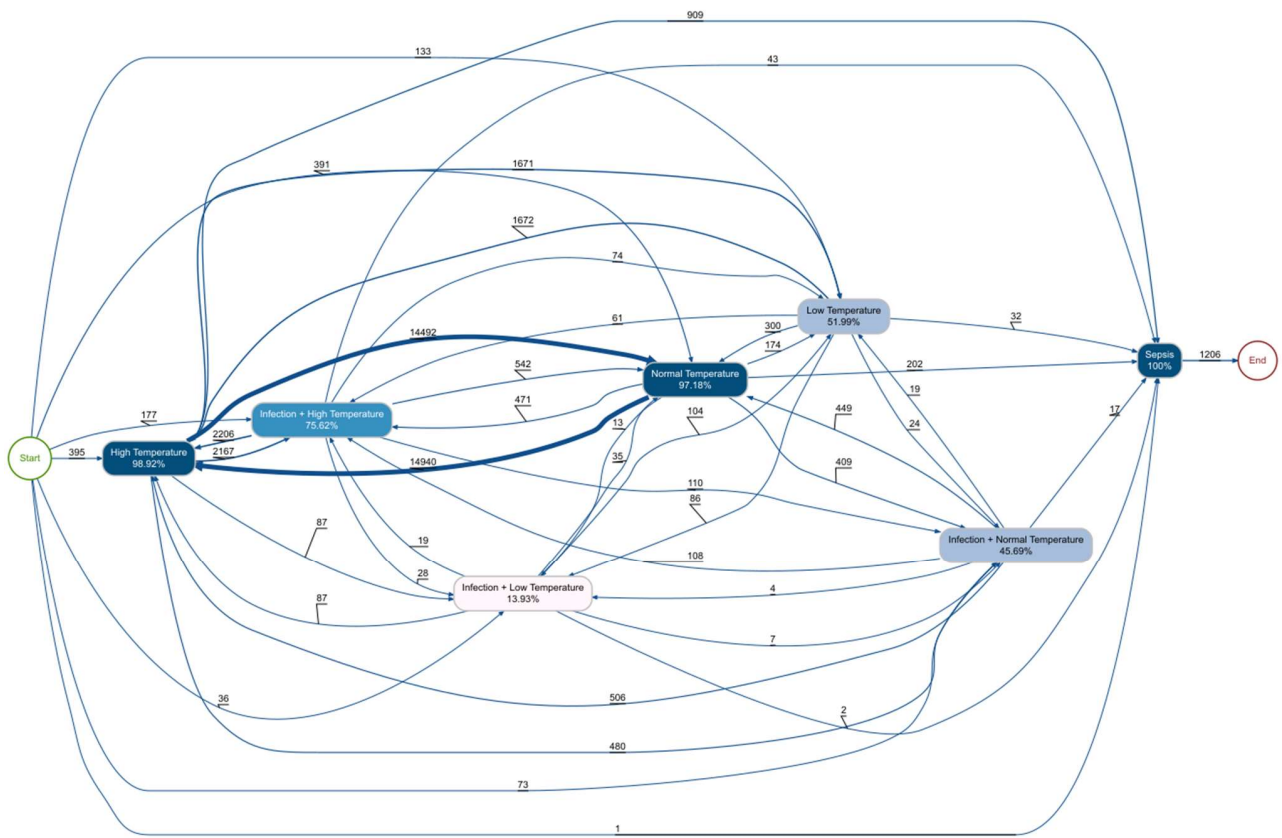

**Figure 5:** Process map of Case I illustrating infection progression. The model depicts transitions between clinical states, highlighting event frequencies and pathways derived from patient data logs.

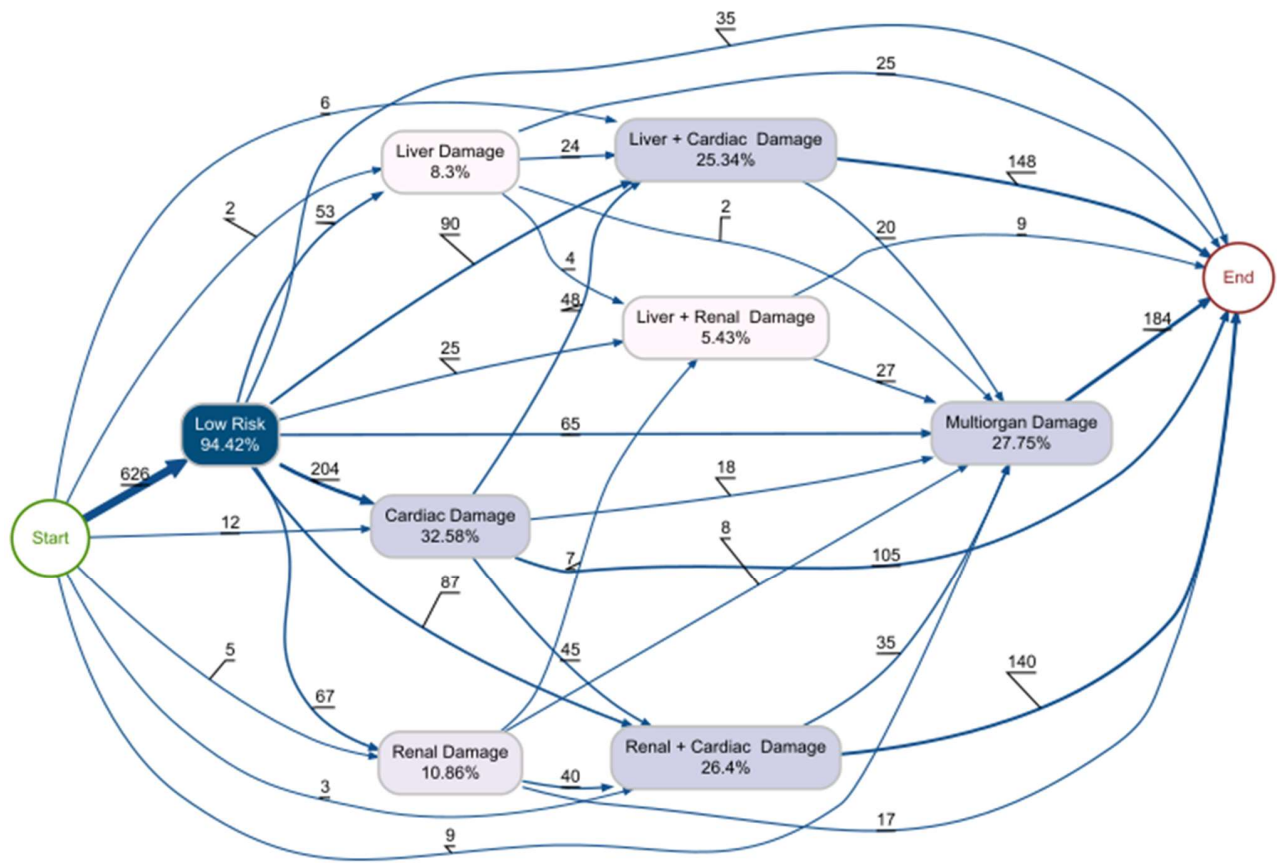

**Figure 6:** Process map for Case II depicting the modeled trajectories of organ damage in the absence of sepsis, with transition probabilities and cumulative outcome distributions

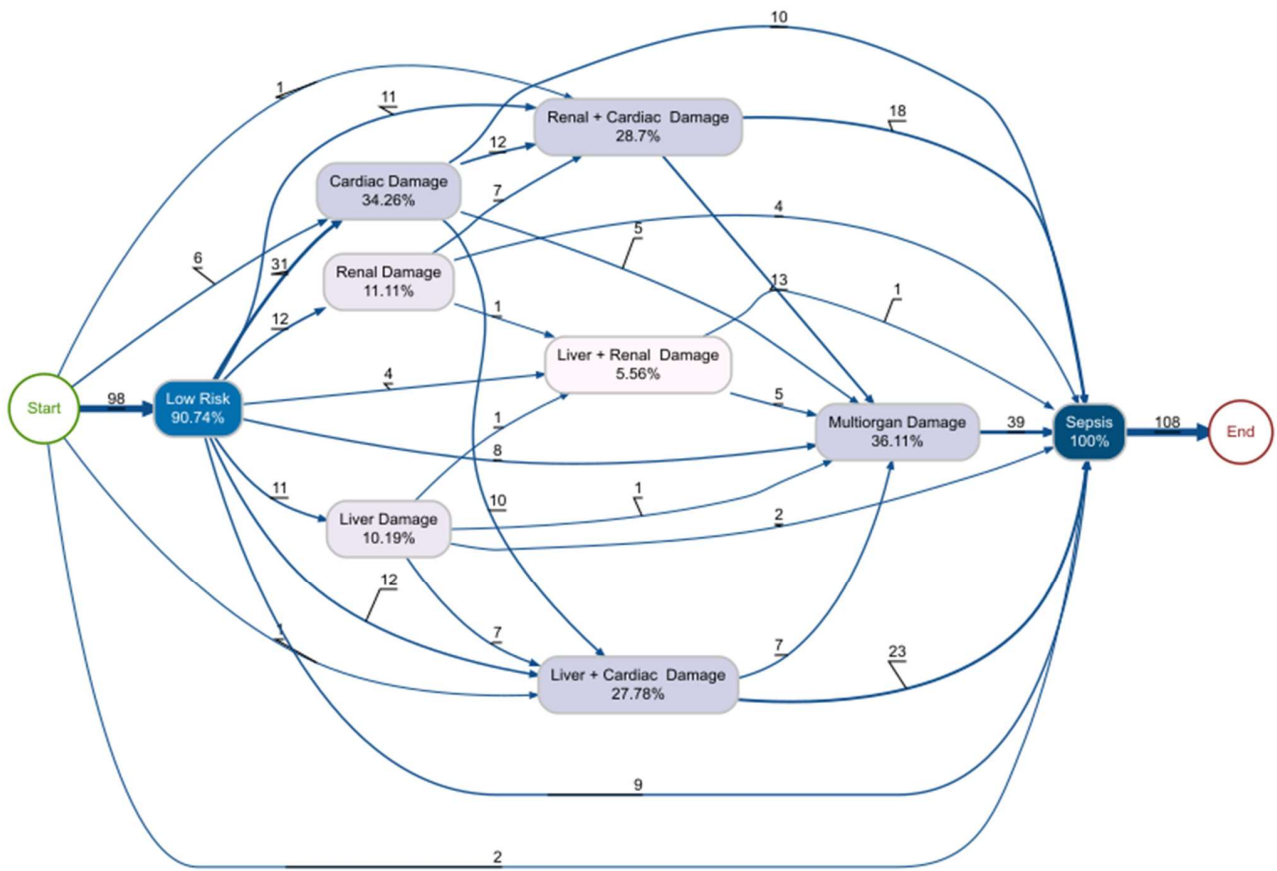

**Figure 7**: Process map for Case II illustrating the modeled trajectories of organ damage in sepsis, with transition probabilities and cumulative outcome distributions.

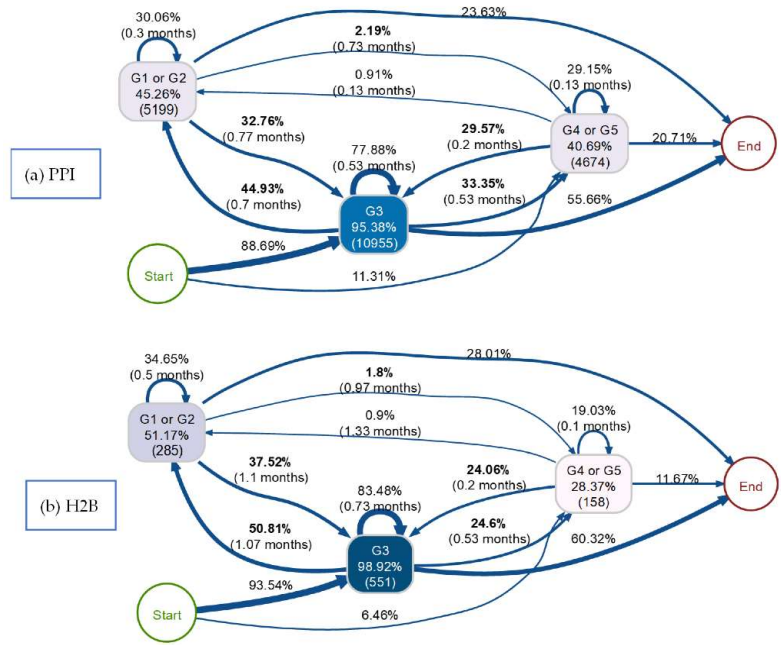

**Figure 8**: Process maps for Case III illustrating the differences between the PPI and H2B groups. Adapted from [24].

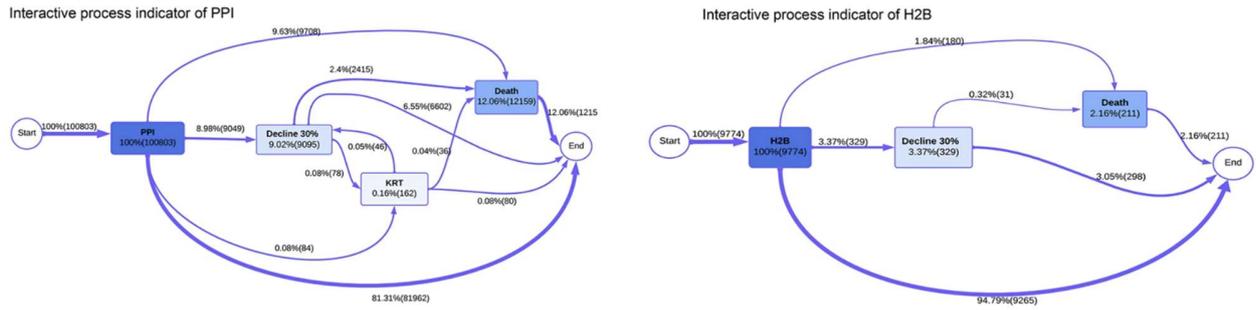

**Figure 9**: Process maps for Case IV illustrating the differences between the PPI group and the H2B group regarding the progression of kidney disease. Adapted from [6].

## Tables

**Table 1:** Specifications and integration characteristics of selected Large Language Models (LLMs) used within the HealthProcessAI framework. The models vary in context window size, token pricing, and domain strengths. Integration is tailored to clinical use cases such as diagnostic reasoning, long-form medical narratives, statistical interpretation, and patient-centered analysis.

| MODEL | PROVIDER | CONTEXT WINDOW | COST TOKENS(*) | PRIMARY STRENGTHS | CLINICAL APPLICATIONS |
|---|---|---|---|---|---|
| **CLAUDE SONNET-4** | Anthropic | 200K tokens | $3.00/$15.00 | Clinical reasoning, guideline interpretation | Complex diagnostic pathways |
| **GPT-4.1** | OpenAI | 128K tokens | $10.00/$30.00 | Broad medical knowledge, consistency | General clinical analysis |
| **GEMINI 2.5 PRO** | Google | 1M tokens | $1.25/$5.00 | Large context, comprehensive analysis | Long clinical narratives |
| **DEEPSEEK R1** | DeepSeek | 64K tokens | $0.55/$2.19 | Technical precision, quantitative analysis | Statistical nterpretation |
| **GROK-4** | X-AI | 128K tokens | $5.00/$15.00 | Creative insights, patient perspectives | Alternative viewpoints |

\* Token pricing is reported per 1 million input/output tokens as of August 2025, with the first price referring to input and the second to output tokens

**Table 2**: Evaluation criteria for assessing LLM-generated clinical reports within the HealthProcessAI framework. Each criterion is assigned a relative weight based on its importance to clinical utility and interpretability. Validation methods involve domain-specific assessments, including clinical expert review, technical accuracy checks, and implementation feasibility testing.

| CRITERION | WEIGHT | DESCRIPTION | VALIDATION METHOD |
|---|---|---|---|
| CLINICAL ACCURACY | 25% | Correctness of medical interpretations and terminology usage | Expert clinical review |
| PROCESS MINING UNDERSTANDING | 20% | Accurate interpretation of analytical results | Technical validation |
| ACTIONABLE INSIGHTS | 20% | Quality and feasibility of clinical recommendations | Implementation assessment |
| STATISTICAL INTERPRETATION | 15% | Correct analysis of quantitative findings | Statistical validation |
| REPORT STRUCTURE & CLARITY | 10% | Organization and readability | Communication analysis |
| EVIDENCE-BASED REASONING | 10% | Use of clinical evidence and literature | Evidence synthesis evaluation |

*Note: Validation methods represent approaches for future clinical studies. Current validation uses automated LLM evaluation.*

**Table 3**: Comparison of process discovery algorithms based on implementation platform, model complexity, processing time, F1-score, and clinical interpretability. Directly-Follows Graph and Inductive Miner are available in both R and Python; others are Python-only.

| ALGORITHM | IMPLEMENTATION | MODEL ELEMENTS | TIME | F1 SCORE | CLINICAL INTERPRETABILITY |
|---|---|---|---|---|---|
| DIRECTLY-FOLLOWS GRAPH | Both platforms | 15 activities, 42transitions | 1.2s | 0.89 | High (4.2/5.0) |
| | | | 2.1s | | |
| HEURISTICS MINER | Python (PM4PY) | 12 places, 15 transitions | 2.8s | 0.85 | High (4.1/5.0) |
| ALPHA ALGORITHM | Python only | 18 places, 22 transitions | 1.9s | 0.76 | Medium (3.4/5.0) |
| INDUCTIVE MINER | Both (varying completeness) | 8 operators | 3.4s | 0.82 | Medium (3.6/5.0) |
| | | | 5.1s | | |
| ILP MINER | Pythons only | 14 places, 19 transitions | 12.3s | 0.79 | Low (2.8/5.0) |

*Note: Each miner generates different graph formats with different elements: activities, places and operators.*

**Table 4**: Performance comparison of LLMs across four proof-of-concept case scenarios: infection, organ dysfunction, glomerular filtration rate (GFR), and kidney outcomes. Scores reflect average ratings on a 4-point scale, with 95% confidence intervals and overall ranking.

| MODEL | CASE I (INFECTION) | CASE II (ORGAN) | CASE III (GFR) | CASE IV (KIDNEY) | OVERALL MEAN | 95% CI | RANK |
|---|---|---|---|---|---|---|---|
| CLAUDE SONNET-4 | 3.75/4.0 | 3.90/4.0 | 3.77/4.0 | 3.85/4.0 | **3.82/4.0** | [3.71-3.93] | 1st |
| GEMINI 2.5 PRO | 3.69/4.0 | 3.77/4.0 | 3.82/4.0 | 3.20/4.0 | **3.61/4.0** | [3.17-4.00] | 2nd |
| GROK-4 | 3.14/4.0 | 3.41/4.0 | 3.12/4.0 | 3.17/4.0 | **3.21/4.0** | [2.44-3.97] | 3rd |
| DEEPSEEK R1 | 3.02/4.0 | 3.30/4.0 | 2.68/4.0 | 3.82/4.0 | **3.18/4.0** | [3.00-3.42] | 4th |
| GPT-4.1 | 3.13/4.0 | 2.72/4.0 | 2.62/4.0 | 3.80/4.0 | **3.04/4.0** | [2.21-3.92] | 5th |

**Table 5**: Cost-effectiveness analysis of LLMs across multi-model evaluations. The table reports average input and output token usage, estimated cost per report, total cost across 20 reports, and the performance-to-cost ratio. DeepSeek R1 demonstrated the highest cost-effectiveness, while GPT-4.1 was the most expensive relative to performance.

| MODEL | INPUT TOKENS | OUTPUT TOKENS | COST | TOTAL COST | PERFORMANCE/COST |
|---|---|---|---|---|---|
| DEEPSEEK R1 | 2,487±142 | 1,234±89 | $0.02 | $0.40 | 154.0 |
| GEMINI 2.5 PRO | 2,523±158 | 1,189±76 | $0.11 | $2.20 | 34.1 |
| CLAUDE SONNET-4 | 2,501±134 | 1,267±103 | $0.26 | $5.20 | 14.7 |
| GROK-4 | 2,489±149 | 1,198±82 | $0.61 | $12.20 | 5.0 |
| GPT-4.1 | 2,476±127 | 1,223±94 | $1.13 | $22.60 | 2.3 |

**Table 6**: Comparative methodological analysis of multi-model AI orchestration applied to healthcare process mining across four proof-of-concept scenarios.

| CASE STUDY | KEY CONSENSUS FINDING | CRTICIAL METRICS | UNIQUE MODEL INSIGHTS | CLINICAL IMPLICATIONS |
|---|---|---|---|---|
| **CASE I** | Temperature fluctuations as central indicators | 6-7h intervention window<br><br>14,940 Normal→High transitions<br><br>>3 cycles = 2-3x sepsis risk | [Anthropic]: 6h window hypothesis<br><br>[Gemini]: Temperature chattering<br><br>[Grok]: Loop frequency model | Early intervention protocols based on temperature volatility |
| **CASE II** | Cardiac damage as gateway to sepsis | 68% of sepsis via cardiac route<br><br>90.7% originate from Low Risk<br><br>57-93h therapeutic window | [Gemini]: "Slow burn" hypothesis<br><br>[Anthropic]: Therapeutic windows<br><br>[DeepSeek]: 3x cardiac risk multiplier | Multi-organ monitoring with cardiac biomarkers |
| **CASE III** | Faster CKD progression with PPI vs H2B | PPI: 9.39 weeks G1/G2→G3<br><br>H2B: 12.09 weeks G1/G2→G3<br><br>20% less time in G3 with PPI | [Anthropic]: Hypomagnesemia pathway<br><br>[Gemini]: Confounding emphasis<br><br>[Grok]: Variant analysis (15% vs 18%) | Enhanced GFR monitoring for PPI patients |
| **CASE IV** | Higher adverse outcomes with PPI in sepsis survivors | 2.7-9x higher eGFR decline risk<br><br>PPI: 9.0% vs H2B: 3.4% major decline<br><br>18–24 month median progression | [Anthropic]: Comprehensive risk framework<br><br>[DeepSeek/Gemini]: Confounding analysis<br>[Grok]: Precise statistics (9% vs 3.4%) | Risk stratification and deprescribing protocols |